\documentclass[conference]{IEEEtran}
\IEEEoverridecommandlockouts
\usepackage{cite}
\usepackage{amsmath,amssymb,amsfonts}
\usepackage{algorithmic}
\usepackage{graphicx}
\usepackage{multirow}
\usepackage{colortbl}
\usepackage{array}
\usepackage{booktabs}
\definecolor{lightblue}{RGB}{220,235,255}
\usepackage{textcomp}
\usepackage{xcolor}
\usepackage{bm}
\def\BibTeX{{\rm B\kern-.05em{\sc i\kern-.025em b}\kern-.08em
    T\kern-.1667em\lower.7ex\hbox{E}\kern-.125emX}}
\begin{document}

\title{HoloEv-Net: Efficient Event-based Action Recognition via Holographic Spatial Embedding and Global Spectral Gating}
\author{\IEEEauthorblockN{Weidong Hao}}

\maketitle

\begin{abstract}
Event-based Action Recognition (EAR) has attracted significant attention due to the high temporal resolution and high dynamic range of event cameras. However, existing methods typically suffer from (i) the computational redundancy of dense voxel representations, (ii) structural redundancy inherent in multi-branch architectures, and (iii) the under-utilization of spectral information in capturing global motion patterns. To address these challenges, we propose an efficient EAR framework named HoloEv-Net. First, to simultaneously tackle representation and structural redundancies, we introduce a Compact Holographic Spatiotemporal Representation (CHSR). Departing from computationally expensive voxel grids, CHSR implicitly embeds horizontal spatial cues into the Time-Height ($\bm{T}$-$\bm{H}$) view, effectively preserving 3D spatiotemporal contexts within a 2D representation. Second, to exploit the neglected spectral cues, we design a Global Spectral Gating (GSG) module. By leveraging the Fast Fourier Transform (FFT) for global token mixing in the frequency domain, GSG enhances the representation capability with negligible parameter overhead. Extensive experiments demonstrate the scalability and effectiveness of our framework. Specifically, HoloEv-Net-Base achieves state-of-the-art performance on THU-EACT-50-CHL, HARDVS and DailyDVS-200, outperforming existing methods by 10.29\%, 1.71\% and 6.25\%, respectively. Furthermore, our lightweight variant, HoloEv-Net-Small, delivers highly competitive accuracy while offering extreme efficiency, reducing parameters by 5.4$\times$, FLOPs by 300$\times$, and latency by 2.4$\times$ compared to heavy baselines, demonstrating its potential for edge deployment.
\end{abstract}


\section{Introduction}
Human Action Recognition (HAR) is pivotal for intelligent systems\cite{carreira2017quo,losey2020controlling,sigurdsson2016hollywood}, yet standard frame-based approaches falter under high-speed or High-Dynamic-Range (HDR) conditions due to motion blur and limited dynamic range. Conversely, bio-inspired event cameras represent a paradigm shift. By asynchronously recording pixel-wise brightness changes, they offer microsecond-level latency and superior HDR ($>$120 dB). These properties make Event-based Action Recognition (EAR) a robust solution for challenging environments where traditional sensors fail.

\begin{figure}[t]  
\centering  
\includegraphics[width=1.0\linewidth]{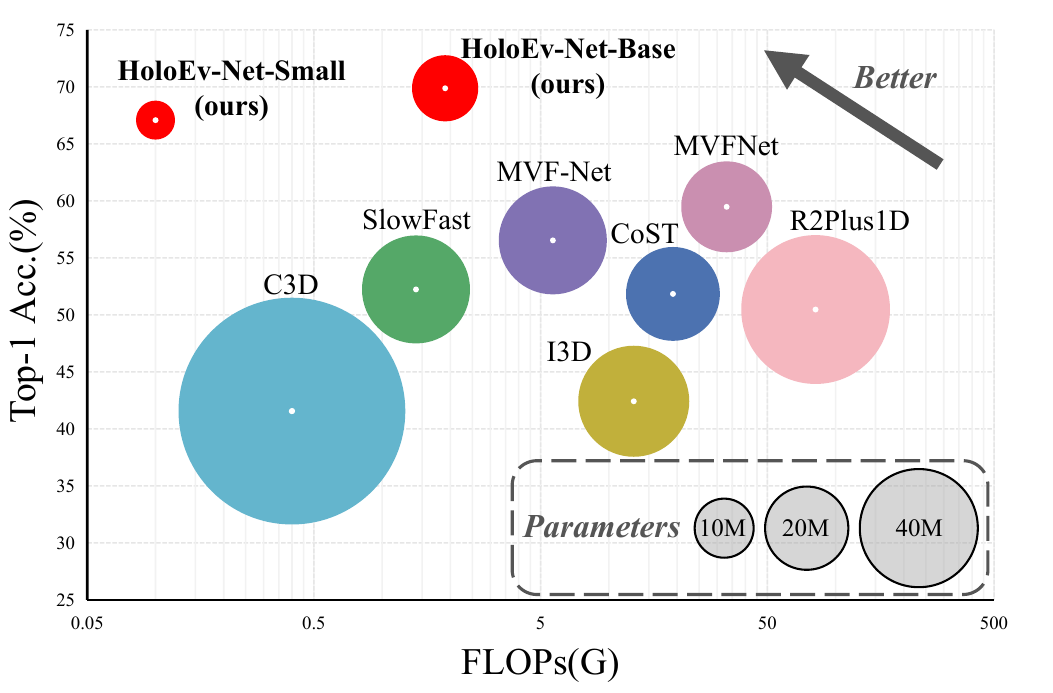}  
\caption{Performance vs. Cost on DailyDVS-200\cite{wang2024dailydvs}. Bubble size denotes model parameters. Our \textbf{HoloEv-Net (red)} outperforms existing methods, demonstrating an optimal balance of accuracy and efficiency.}  
\label{fig:acc_compare}  
\end{figure}  

Current state-of-the-art EAR methods\cite{wang2024dailydvs,wang2024hardvs,gao2023action} largely rely on voxel representations. While effective, transforming sparse event streams into dense 3D spatiotemporal bins often introduces \textbf{computational redundancy}\cite{fan2025eventpillars}. Processing such high-dimensional inputs necessitates heavy backbones, contradicting the low-latency and energy-efficient nature of neuromorphic sensors. To alleviate this computational burden, multi-view learning, which is a paradigm widely adopted in HAR\cite{li2019collaborative,wu2021mvfnet} fuse temporal and spatial dimensions via distinct projections (e.g., Height-Width ($H$-$W$), Time-Width ($T$-$W$), Time-Height ($T$-$H$) views). However, this approach introduces a new bottleneck: \textbf{structural redundancy}. Standard multi-view frameworks typically rely on multi-branch architectures, where each view requires a dedicated backbone for feature extraction, leading to significant parameter overhead and computational costs. 

To resolve above challenges, we propose the \textbf{C}ompact \textbf{H}olographic \textbf{S}patiotemporal \textbf{R}epresentation (\textbf{CHSR}). Unlike standard multi-view paradigms that require parallel branches, CHSR achieves multi-view perception within a single view by holographically embedding horizontal spatial cues into the $T$-$H$ plane. This strategy yields a compact 2D representation that preserves continuous temporal dynamics and 3D spatial contexts, thereby simultaneously alleviating the computational redundancy of voxel grids and the structural redundancy of multi-branch architectures.

Beyond redundancy, existing frameworks face another limitation: \textbf{under-utilization of spectral information}. Human actions inherently possess global periodic patterns\cite{runia2018real} that are difficult for lightweight CNNs with limited local receptive fields to capture\cite{rao2021global}. While our CHSR effectively transforms these temporal periodicities into distinct wave-like textures within the $T$-$H$ view (see Fig.~\ref{fig:freq_analysis}), standard spatial convolutions are ill-equipped to analyze such global spectral properties.

To address this, we introduce the \textbf{G}lobal \textbf{S}pectral \textbf{G}ating (GSG) module. Capitalizing on the explicit spectral textures exposed by CHSR, GSG leverages the Fast Fourier Transform (FFT) to perform global token mixing in the frequency domain. This mechanism allows the network to efficiently capture long-range dependencies and intrinsic motion periodicities with negligible parameter overhead, perfectly complementing the local operations of the lightweight backbone.

\begin{figure}[t]  
\centering  
\includegraphics[width=1.0\linewidth]{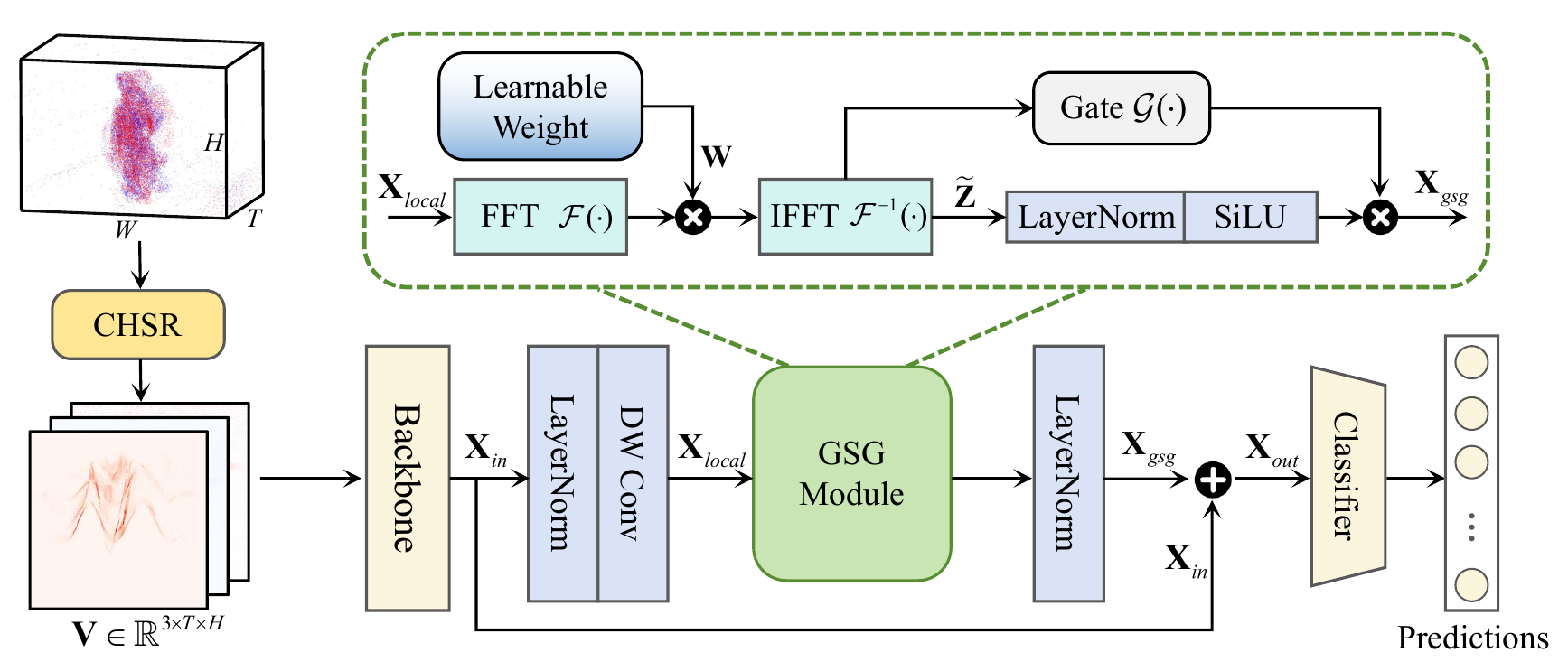}  
\caption{Overview of the proposed HoloEv-Net. The raw event stream is first processed by our Compact Holographic Spatiotemporal Representation (CHSR). The extracted features from the backbone are processed by the Global Spectral Gating (GSG) module.}
\label{fig:model_structure}  
\end{figure}

We term this efficient framework \textbf{HoloEv-Net}. By integrating the spatially-aware CHSR with the frequency-domain GSG module, our architecture effectively harmonizes representation completeness with model efficiency (Fig.~\ref{fig:model_structure}). To demonstrate its scalability, we instantiate two variants: a high-performance HoloEv-Net-Base and a lightweight HoloEv-Net-Small. Our main contributions are summarized as follows:
\begin{itemize}
\item We propose the Compact Holographic Spatiotemporal Representation (CHSR) to simultaneously dismantle the computational and structural redundancies of existing voxel-based or multi-view paradigms.
\item We design a Global Spectral Gating (GSG) module to address CNNs’ deficiency in modeling global motion periodicities. Employing FFT in the frequency domain, GSG strengthens feature representation and complements local spatial features with negligible parameter overhead.
\item Extensive experiments demonstrate the superiority of our framework. HoloEv-Net-Base achieves SOTA performance on THU-EACT-50-CHL\cite{gao2023action}, DailyDVS-200\cite{wang2024dailydvs}, and HARDVS\cite{wang2024hardvs}. Meanwhile, HoloEv-Net-Small maintains competitive accuracy while significantly reducing parameters, FLOPs and latency by 5.4$\times$, 300$\times$ and 2.4$\times$.
\end{itemize}

\section{Related Work}

\subsection{Dense Event Representations}
Event cameras generate sparse, asynchronous streams, rendering them incompatible with standard CNNs. Thus, constructing dense representations is a prerequisite for EAR. Early methods\cite{maqueda2018event,cladera2020device} aggregated sparse event data along the temporal axis to encode them into 2D images. While bridging the gap to frame-based algorithms, this inevitably compresses rich temporal cues. Furthermore, the voxel grid\cite{zhu2018ev} preserves temporal cues by partitioning the event stream into spatiotemporal bins. This paradigm currently dominates EAR\cite{wang2024dailydvs,wang2024hardvs,wang2024event,gao2023action}, as dense representations within time windows effectively capture motion evolution. However, processing such high-dimensional structures inherently introduces significant computational redundancy.

\subsection{Multi-View Learning in Action Recognition}
To solve this problem, multi-view learning improves action recognition by exploiting complementary spatiotemporal cues from orthogonal projections\cite{li2019collaborative,wu2021mvfnet}. However, this paradigm typically relies on multi-branch architectures, where each view requires a dedicated backbone, leading to significant parameter overhead. Furthermore, aggregating these views necessitates complex fusion strategies, ranging from simple concatenation\cite{deng2021mvf} to intricate attention mechanisms\cite{song2024gaf} or temporal sequence modeling\cite{ullah2021conflux}. These explicit fusion designs are often computationally expensive and difficult to optimize. Unlike these heavy architectures, we demonstrate that multi-view perception can be effectively achieved within a single lightweight branch via holographic spatial embedding, eliminating the need for elaborate fusion modules.

\subsection{Frequency Domain Learning in Vision}

Frequency analysis is a powerful tool for capturing global dependencies\cite{FTC-net,rao2021global}. In Human Activity Recognition (HAR), FTC-Net\cite{FTC-net} demonstrated that repetitive activities (e.g., walking, running) exhibit distinct spectral signatures, where Fourier Transform Convolutions effectively extract these intrinsic periodicities. For event-based vision, the sensor's microsecond-level temporal resolution makes it ideal for frequency analysis. Recent biological studies\cite{hamann2025fourier} have validated that high-frequency periodic behaviors in wildlife can be precisely quantified via Fast Fourier Transform (FFT), proving that spectral features often possess greater discriminative power than complex temporal ones. However, mainstream EAR methods largely overlook global frequency information, relying instead on local spatial convolutions. We argue that frequency analysis naturally aligns with our proposed CHSR. Consequently, we introduce a GSG module to exploit these spectral cues for robust action recognition.

\section{Proposed Method}
\subsection{Overview}
This section outlines our proposed framework, with its overall architecture illustrated in Fig.~\ref{fig:model_structure}. The pipeline begins with the raw event stream, which is processed by our \textbf{C}ompact \textbf{H}olographic \textbf{S}patiotemporal \textbf{R}epresentation (CHSR). As shown on the left of Fig.~\ref{fig:model_structure}, CHSR transforms the sparse, asynchronous event stream into a dense, three-channel spatiotemporal tensor, effectively preserving 3D motion cues within a 2D representation. Subsequently, these encoded feature maps are fed into a lightweight backbone. To capture long-range dependencies that are typically missed by local convolutions, we integrate our novel \textbf{G}lobal \textbf{S}pectral \textbf{G}ating (\textbf{GSG}) module. As depicted in the expanded view (top dashed box), the GSG module leverages Fast Fourier Transform (FFT) to perform global token mixing in the frequency domain. Finally, the refined features pass through a classifier for action prediction. We elaborate on the details of these modules in the following subsections.

\begin{figure}[t]  
\centering  
\includegraphics[width=0.95\linewidth]{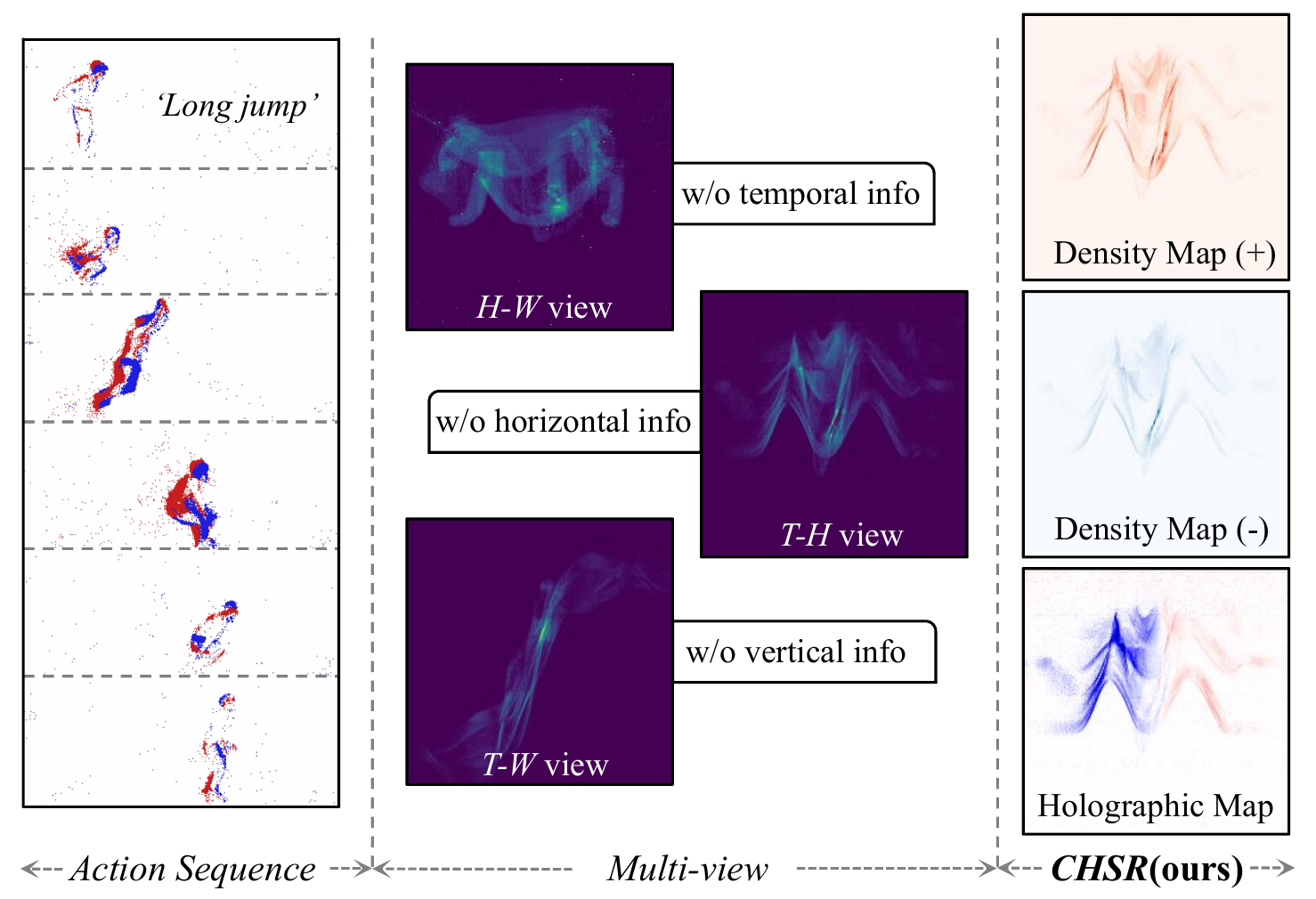}  
\caption{Visualization of the proposed CHSR. Unlike standard multi-view projections (middle) that suffer from information loss, our CHSR constructs a comprehensive representation in the $T$-$H$ domain. It consists of three channels: \textbf{(1) Density Map (+)} and \textbf{(2) Density Map (-)}, which accumulate polarity-specific events to record motion trajectories; and \textbf{(3) Holographic Map}, which recovers the horizontal spatial cues (different color denotes different horizontal position) typically lost in $T$-$H$ projections.}  
\label{fig:representation}  
\end{figure}  

\begin{figure*}[t]  
\centering  
\includegraphics[width=0.8\linewidth]{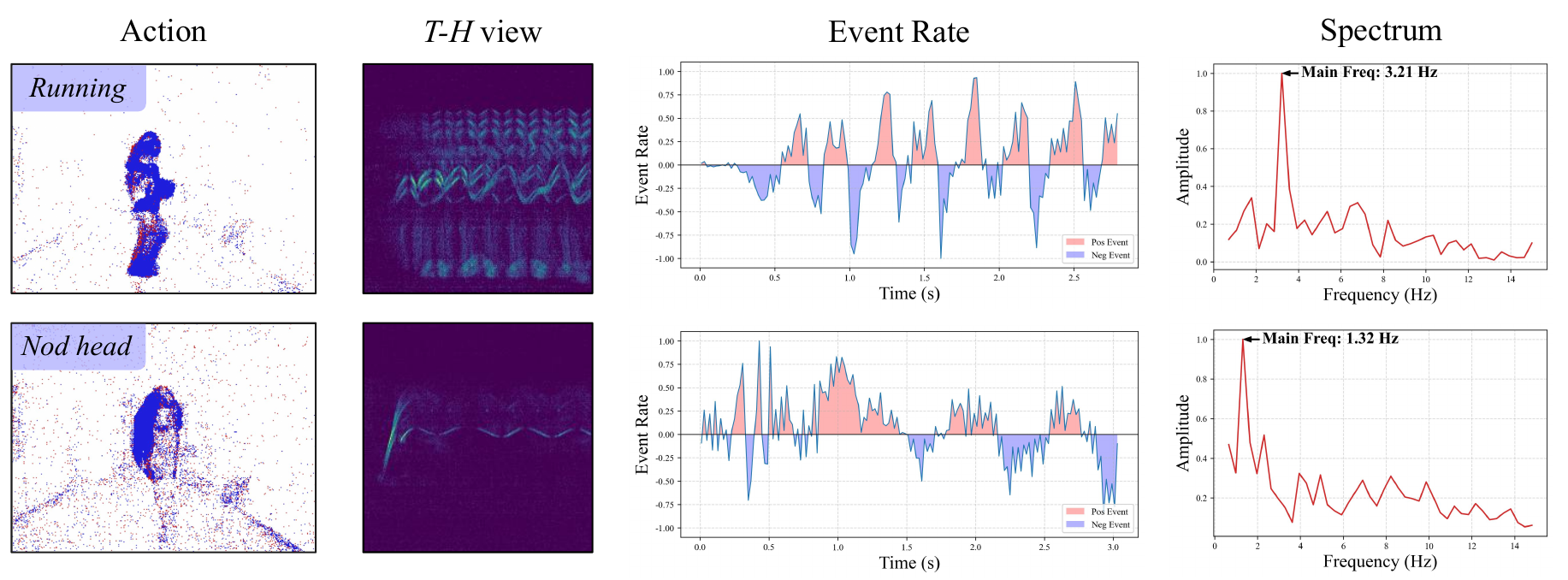}  
\caption{Frequency analysis of event streams. The figure displays the event rate and FFT spectrum for two distinct actions. The distinct main frequencies (3.21 Hz for running, 1.32 Hz for nodding) highlight the discriminability of actions in the frequency domain.}  
\label{fig:freq_analysis}  
\end{figure*}

\subsection{CHSR Representation}
\paragraph{Analysis}
Current representation methods for EAR tasks primarily rely on voxel grid, which encode the event stream into multiple event frames. Thus, the spatiotemporal cues are embedded between frames. Although simply encoding the event stream into $H$-$W$, $T$-$W$, and $T$-$H$ views can directly embed spatiotemporal cues within frames, it necessitates complex multi-branch architectures and fusion strategies, where selecting the fusion stage is notably difficult. To this end, we rethink whether all three views are essential for EAR and investigate if complete spatiotemporal cues can be encoded within a single view.
\paragraph{Implementation}
Among the three views of encoded event streams, the $H$-$W$ view lacks the temporal cues\cite{zhu2018ev}, which is crucial for EAR tasks. Between the temporal views ($T$-$W$ and $T$-$H$), the $T$-$H$ view is superior as it preserves the vertical topological structure of the human body, directly facilitating action classification. Conversely, the $T$-$W$ view provides mainly auxiliary information, such as horizontal positioning. As illustrated in Fig.~\ref{fig:representation} (middle), standard projections inevitably suffer from dimensional information loss (e.g., the $T$-$H$ view loses horizontal context). Therefore, we embed horizontal information, which comes from $T$-$W$ view, into the $T$-$H$ view to form a comprehensive representation. Detailed ablation studies validating this view selection are presented in the next section (Tab.~\ref{tab:ablation_view}). Formally, given an input event stream:
\begin{align}  
\label{eq:event_def} 
\mathcal{E} = \{e_k\}_{k=1}^N = \{(x_k,y_k,t_k,p_k)\}_{k=1}^N,
\end{align}
where $(x_k,y_k)$ denotes the spatial coordinates, and $t_k$ represents the timestamp. The polarity $p_k \in \{-1,1\}$ indicates the sign of brightness change, dividing the stream into positive events ($\mathcal{E}_+$) and negative events ($\mathcal{E}_-$).
we construct a dense tensor $\mathbf{V} \in \mathbb{R}^{3 \times T \times H}$. The tensor $\mathbf{V}$ is composed of three complementary feature maps: positive density map ($\mathcal{D}^+$), negative density map ($\mathcal{D^-}$) and holographic map ($\mathcal{H}$), as visualized in the right column of Fig.~\ref{fig:representation}.
\begin{align}
\mathcal{D}^\pm(t, y) \doteq \sum_{e_k \in \mathcal{E}_\pm} \delta(t - t_k, y - y_k),
\end{align}
where $\delta$ represents the Kronecker delta function. By aggregating the impulses based on event polarity, this function effectively accumulates the local event density over time in the $T$-$H$ view.

\begin{align} \mathcal{H}(t, y) \doteq \sum_{k=1}^{N} \phi(x_k) \cdot \delta(t - t_k, y - y_k), \end{align}
where $\phi(x_k) = sin(\pi x_k / W)$ denotes the transverse spatial embedding function. By transforming horizontal coordinates into cyclic sinusoidal patterns, this operation implicitly preserves essential 3D spatiotemporal contexts within the 2D $T$-$H$ view, effectively realizing a holographic encoding mechanism.

\begin{table*}[!t]
\centering
\caption{\textbf{Comparisons on THU-EACT-50-CHL, HARDVS and DailyDVS-200.} Best results in \textbf{bold} and second best \underline{underlined}.}
\label{tab:acc_comparison}
\scriptsize
\setlength{\tabcolsep}{2pt} 
\resizebox{\linewidth}{!}{ 
\begin{tabular}{lcc *{6}{>{\centering\arraybackslash}p{1cm}} ccc}
\toprule
\multirow{2}{*}{\textbf{Methods}} & \multirow{2}{*}{\textbf{Input Type}} & \multirow{2}{*}{\textbf{Backbone}} & \multicolumn{2}{c}{\textbf{THU-EACT-50-CHL}} & \multicolumn{2}{c}{\textbf{HARDVS}} & \multicolumn{2}{c}{\textbf{DailyDVS-200}} & \multirow{2}{*}{\textbf{FLOPs}} & \multirow{2}{*}{\textbf{Params}} & \multirow{2}{*}{\textbf{Latency}}\\

& & & Top-1(\%) & Top-5(\%) & Top-1(\%) & Top-5(\%) & Top-1(\%) & Top-5(\%) & & &\\
\midrule
R2Plus1D~\cite{tran2018closer} & \multirow{5}{*}{Frame}  & ResNet34 & 50.54 & 73.15 & 49.06 & 56.43 & 36.06 & 63.67 & 81.4G & 63.5M &11.2ms\\
SlowFast~\cite{feichtenhofer2019slowfast} & & ResNet50 & 46.14 & 73.16 & 50.63 & 57.77 & 41.49 & 68.19 & 1.4G & 33.6M &19.6ms\\
TSM~\cite{lin2019tsm} & & ResNet50 & 44.30 & 71.31 & 52.63 & 60.56 & 40.87 & 71.46 & 1.4G & 24.3M &11.8ms\\
I3D~\cite{carreira2017quo} & & ResNet50 & 42.41 & 68.49 & 51.53 & 59.82 & 32.30 & 59.05 & 12.8G & 35.3M &13.1ms\\ 
C3D~\cite{tran2015learning} & & 3D CNN & 41.54 & 68.57 & 50.52 & 56.14 & 21.99 & 45.81 & \underline{0.4G} & 147.2M &\underline{8.4ms}\\
\midrule
EST~\cite{gehrig2019end} & Learnable & ResNet34 & 55.07 & 76.02 & 36.51 & 42.09 & 32.23 & 59.66 & 4.2G & 21.5M &11.7ms\\
\midrule
CoST~\cite{li2019collaborative} & \multirow{3}{*}{Multi-view} & ResNet50 & 51.84 & 75.00 & 50.57 & 61.38 & 36.09 & 64.45 & 19.2G & 25.4M &21.3ms\\
MVFNet~\cite{wu2021mvfnet} & & ResNet50 & 59.56 & 77.49 & \underline{52.98} & \underline{63.26} & 48.30 & 75.91 & 32.8G & 23.6M &17.5ms\\
MVF-Net~\cite{deng2021mvf} & & ResNet34\&18 & 56.47 & 76.53 & 52.61 & 61.67 & 43.98 & 70.39 & 5.6G & 33.6M &13.7ms\\
\midrule
\textbf{HoloEv-Net-Base} & \multirow{2}{*}{Holographic} & ResNet-18 & \textbf{69.85} & \textbf{79.60} & \textbf{54.69} & \textbf{63.56} & \textbf{54.55} & \textbf{77.50} & 1.9G & \underline{12.5M} &9.6ms\\
\textbf{HoloEv-Net-Small} & & MobileNetV3-Small & \underline{67.10} & \underline{78.86} & 52.69 & 62.56 & \underline{49.13} & \underline{73.56} & \textbf{0.1G} & \textbf{4.4M} &\textbf{7.4ms}\\
\bottomrule
\end{tabular}
} 
\end{table*}
\subsection{GSG Module}
\paragraph{Analysis}
As illustrated in Fig.~\ref{fig:freq_analysis}, human actions exhibit distinct periodic oscillatory patterns in the $T$-$H$ view, implying high discriminability in the frequency domain. While standard CNNs excel at extracting local textures, their limited receptive fields struggle to capture these global periodic dependencies. To address this, we propose the Global Spectral Gating (GSG) module, which performs efficient global token mixing in the frequency domain.
\paragraph{Architecture Overview}
As shown in Fig.~\ref{fig:model_structure}, our framework is instantiated with either a ResNet-18\cite{he2016deep} or a MobileNetV3-Small\cite{howard2019searching} backbone to process the CHSR tensor $\mathbf{V} \in \mathbb{R}^{3 \times T \times H}$. The backbone extracts high-level features, denoted as $\mathbf{X}_{in} \in \mathbb{R}^{C \times T' \times H'}$. We integrate the GSG module at the end of the backbone via a global residual connection. This design fuses the local spatial features from the backbone with the spectrally modulated global context, formulated as $\mathbf{X}_{out} = \mathbf{X}_{in} + \mathbf{X}_{gsg}$. This residual structure ensures robust feature learning and stable gradient flow.
\paragraph{Implementation}
The GSG module operates through three sequential stages: \textbf{(1) Local Embedding.} We first apply a $3\times3$ Depthwise Convolution (DWConv) to $\mathbf{X}_{in}$. This step encodes local spatial contexts and prevents high-frequency spectral aliasing.\textbf{(2) Global Spectral Filtering.} To capture long-range dependencies, the features are transformed into the frequency domain via a 2D Real-to-Complex FFT along the temporal ($T$) and height ($H$) axes. A learnable complex weight tensor $\mathbf{W}$ modulates the spectrum to emphasize action-specific frequencies:
\begin{align}
\label{eq:spectral_mix}\tilde{\mathbf{Z}} = \mathcal{F}^{-1}\left( \mathcal{F}(\mathbf{X}_{local}) \odot \mathbf{W} \right),
\end{align}
where $\mathcal{F}(\cdot)$/$\mathcal{F}^{-1}(\cdot)$ denote FFT/Inverse FFT. This operation achieves global information interaction.\textbf{(3) Gated Reconstruction.} To suppress spectral artifacts, the refined features $\tilde{\mathbf{Z}}$ pass through a dual-branch gating mechanism. One branch acts as a content carrier (normalized by LayerNorm and activated by SiLU), while the other serves as a dynamic gate $\mathcal{G}(\cdot)$:
\begin{align}\label{eq:gating}\mathbf{X}_{gsg} = \text{SiLU}(\text{LN}(\tilde{\mathbf{Z}})) \odot \mathcal{G}(\tilde{\mathbf{Z}}).
\end{align}
This gating mechanism selectively filters noise while retaining salient motion features before the final residual fusion.

\section{Experiment}
\subsection{Implementation Details}
\paragraph{Datasets} To evaluate the effectiveness and robustness of our proposed method, we conduct comprehensive experiments on three challenging event-based action recognition benchmarks: THU-EACT-50-CHL\cite{gao2023action}, HARDVS\cite{wang2024hardvs} and DailyDVS-200\cite{wang2024dailydvs}. THU-EACT-50-CHL contains 2,330 samples across 50 categories, featuring distinct lighting conditions. HARDVS is a large-scale dataset comprising 107,646 sequences of 300 action classes performed by 5 subjects. DailyDVS-200 includes 22,046 sequences covering 200 daily action categories, characterized by 14 environmental challenges such as camera motion and illumination changes.
\paragraph{Training Protocol} All experiments were implemented using the PyTorch framework and conducted on a single NVIDIA RTX 3090 GPU. Following\cite{deng2021mvf}, the $T$ axis is discretized into $T$ = 224. The network was trained for 50 epochs using the Adam optimizer with an initial learning rate of $1e-4$. The batch size is set to 32. We used the Cross-Entropy loss function for optimization.
\paragraph{Evaluation Metrics} To measure recognition accuracy, we report Top-1 and Top-5 accuracy. Beyond accuracy, to assess the model's complexity and computational efficiency, we provide model parameters (Params) and floating-point operations (FLOPs). Furthermore, we measure the inference Latency (ms/sample) to evaluate suitability for edge deployment.

\subsection{Quantitative Results}
We compare our HoloEv-Net with recent state-of-the-art approaches, including frame-based\cite{tran2018closer,feichtenhofer2019slowfast,lin2019tsm,carreira2017quo,tran2015learning}, learnable\cite{gehrig2019end}, and multi-view\cite{li2019collaborative,wu2021mvfnet,deng2021mvf} methods. The quantitative results on THU-EACT-50-CHL, HARDVS and DailyDVS-200 are summarized in Tab.~\ref{tab:acc_comparison}.
\paragraph{Performance Improvement} Our HoloEv-Net-Base achieves 69.85\%, 54.69\% and 54.55\% Top-1 accuracy on three datasets, outperforming the leading method MVFNet\cite{wu2021mvfnet} by 10.29\%, 1.71\% and 6.25\%, respectively.
\paragraph{Efficiency Analysis} In terms of computational cost, our HoloEv-Net-Base requires only 12.5M parameters and 1.86G FLOPs (measured on DailyDVS-200). Furthermore, our HoloEv-Net-Small pushes the limits of efficiency with only 4.4M parameters and 0.1G FLOPs. Crucially, as shown in Table~\ref{tab:acc_comparison}, HoloEv-Net-Small achieves a remarkable latency of 7.4 ms on RTX 3090. More importantly, on the embedded NVIDIA Jetson AGX Orin, it sustains a real-time latency of 32.4 ms, confirming its suitability for edge deployment.

\begin{table}[!t]
\centering
\caption{Ablation study on view selection and embedding strategies on THU-EACT-50-CHL.}
\label{tab:ablation_view}
\small
\setlength{\tabcolsep}{4pt}
\begin{tabular}{llcc}
\toprule
\textbf{Group} & \textbf{View Combination} & \textbf{Top-1 (\%)} & \textbf{FLOPs (G)} \\
\midrule
\multirow{3}{*}{Single-view} 
& $H$-$W$ & 38.60 & 1.82 \\
& $T$-$W$ & 50.18 & 1.82 \\
& $T$-$H$ & 60.36 & 1.82 \\
\midrule
\multirow{4}{*}{Multi-view} 
& $H$-$W$ + $T$-$H$ & 63.91 & 1.90 \\
& $H$-$W$ + $T$-$W$ & 54.23 & 1.90 \\
& $T$-$W$ + $T$-$H$ & 65.07 & 1.90 \\
& $H$-$W$ + $T$-$H$ + $T$-$W$ & \underline{65.09} & 1.98 \\
\midrule
\multirow{2}{*}{Holographic} 
& $H$-$W$ with $H$ & 60.16 & 1.86 \\
& $T$-$H$ with $W$ & \textbf{69.85} & 1.86 \\
\bottomrule
\end{tabular}
\end{table}

\begin{table}[!t]
\centering
\caption{Ablation study on the GSG components on THU-EACT-50-CHL.}
\label{tab:ablation_gsg}
\setlength{\tabcolsep}{10pt} 
\begin{tabular}{lcccc}
\toprule
\textbf{Component} & \multicolumn{4}{c}{\textbf{Variants}} \\
\midrule
Baseline (ResNet)       & \checkmark & \checkmark & \checkmark & \checkmark \\
FFT                     &            & \checkmark & \checkmark & \checkmark \\
Learnable Weight        &            &            & \checkmark & \checkmark \\
Gated Reconstruction    &            &            &            & \checkmark \\
\midrule
\textbf{Top-1 Acc (\%)} & 65.07      & 66.36      & \underline{68.19}       & \textbf{69.85} \\
\bottomrule
\end{tabular}
\end{table}

\subsection{Ablation Study}
\paragraph{Analysis on Representation Construction} We investigate the effectiveness of different projection views and encoding strategies. As shown in Tab.~\ref{tab:ablation_view}, we compare three settings: standard single-view projections, multi-view fusion, and our proposed holographic representation. To ensure fair comparison, the input layer of the backbone is adjusted to accommodate the varying dimensions of different representations. Among single-view, the $H$-$W$ view yields the lowest accuracy (38.60\%). This corroborates that the absence of the temporal dimension leads to a critical loss of motion cues essential for EAR. Furthermore, the $T$-$H$ view consistently outperforms the $T$-$W$ view(+10.18\%). While standard multi-view methods (similar to MVF-Net\cite{deng2021mvf}) attempt to combine information, they suffer from dimensional misalignment—directly concatenating features from heterogeneous domains (e.g. spatial $H$ vs. temporal $T$) introduces optimization difficulties and limits performance gains. Consequently, our Holographic encoding ($T$-$H$ with $W$) proves to be the optimal solution. By embedding the horizontal motion cues into the $T$-$H$ view, it not only achieves the effective coupling of multi-view spatiotemporal information but also circumvents the dimensional mismatch issue in traditional fusion strategies.

\paragraph{Analysis on GSG} We first evaluate the internal components of the GSG module. As summarized in Tab.~\ref{tab:ablation_gsg}, simply introducing the FFT yields a performance gain of 1.29\% (from 65.07\% to 66.36\%) over the baseline. A more significant improvement (+1.83\%) is observed when the weights are made learnable. Finally, integrating the gated reconstruction mechanism further boosts the Top-1 accuracy to 69.85\%. Beyond component analysis, the GSG module proves critical for dynamic environments. In scenarios characterized by significant camera motion, where background noise typically degrades performance, our method demonstrates exceptional stability. As illustrated in Tab.~\ref{tab:motion_robustness}, under the camera motion condition, HoloEv-Net-Base with GSG surpasses Swin-T and HoloEv-Net-Base without GSG by 10.42\% and 8.95\%, respectively. This result confirms that our GSG mechanism effectively filters out motion-induced noise while preserving salient action features.

\begin{table}[t]
\centering
\caption{Analysis against camera motion on DailyDVS-200.}
\label{tab:motion_robustness}
\resizebox{1.0\columnwidth}{!}{
\begin{tabular}{lcc}
    \toprule
    \multirow{2}{*}{\textbf{Method}} & \multicolumn{2}{c}{\textbf{Camera Movement}}\\
    & Static (\%) & Motion (\%)\\
    \midrule
    SlowFast~\cite{feichtenhofer2019slowfast} & 51.81 & 21.05\\
    TSM~\cite{lin2019tsm} & 50.16 & 26.88\\
    EST~\cite{gehrig2019end} & 41.51 & 13.52\\
    Spikeformer~\cite{zhou2022spikformer} & 46.15 & 13.74\\
    Swin-T~\cite{liu2022video} & \underline{58.05} & \underline{27.84}\\
    \midrule
    \textbf{HoloEv-Net-Base (w/o GSG)} & \textbf{58.59} & \textbf{29.31}\\
    \textbf{HoloEv-Net-Base} & \textbf{59.62} & \textbf{38.26}\\
    \bottomrule
\end{tabular}
    }
\end{table}

\begin{table}[!t]
\centering
\caption{Ablation study on the integration stage of the GSG module within the ResNet-18 backbone. }
\label{tab:ablation_stage}
\small
\setlength{\tabcolsep}{10pt}
\resizebox{1.0\columnwidth}{!}{
\begin{tabular}{lcc}
\toprule
\textbf{Insertion Stage} & \textbf{Feature Resolution} & \textbf{Top-1 Acc (\%)} \\
\midrule
Baseline (None)       & -              & 65.07 \\
Stage 1               & $56 \times 56$ & 66.54 \\
Stage 2               & $28 \times 28$ & 66.72 \\
Stage 3               & $14 \times 14$ & \underline{67.83} \\
\textbf{Stage 4 (Ours)} & $\mathbf{7 \times 7}$ & \textbf{69.85} \\
\bottomrule
\end{tabular}
}
\end{table}
\paragraph{Analysis on Module Integration Stages}
Finally, we investigate the optimal position for the GSG module by inserting it after each stage of the ResNet-18 backbone. The results are reported in Tab.~\ref{tab:ablation_stage}. The feature resolution indicates the feature map size at each stage. We observe that the performance improvement becomes more significant as the module is inserted into deeper layers. Inserting GSG at Stage 4 yields the highest Top-1 accuracy of 69.85\%. In the early stages, feature maps possess high spatial resolution and primarily encode low-level visual primitives. Applying global spectral filtering here may inadvertently blur these fine-grained details or be dominated by high-frequency noise. However, in deep stage, the global receptive field provided by the GSG module effectively captures long-range dependencies and the overall temporal rhythm of the action, which are crucial for classification. Moreover, processing lower-resolution feature maps at Stage 4 ($7 \times 7$) is also more computationally efficient.

\section{Conclusion}
In this paper, we proposed efficient HoloEv-Net for EAR task. To address computational and structural redundancy, we introduced CHSR. This strategy implicitly achieves multi-view perception within a single $T$-$H$ view by holographically embedding horizontal spatial cues into the channel dimension. Furthermore, to utilize spectral information of EAR, we designed a Global Spectral Gating (GSG) module, which leverages the FFT to capture long-range dependencies with negligible parameter overhead. Extensive experiments on three benchmark datasets demonstrate the superiority of our framework. HoloEv-Net-Base achieves SOTA accuracy. Meanwhile, HoloEv-Net-Small, pushes the limits of efficiency, achieving low latency suitable for real-time edge deployment.

While achieving SOTA, the modest gains on HARDVS highlight the limitations of our spectral gating in complex scenarios. Future work will focus on adaptive frequency learning to dynamically accommodate diverse motion patterns. Furthermore, we aim to generalize the holographic paradigm to downstream tasks such as object detection and optical flow.

\bibliographystyle{IEEEtran}
\bibliography{IEEEabrv,ref}
\end{document}